\documentclass[runningheads]{llncs}
\usepackage[T1]{fontenc}
\usepackage{graphicx}
\usepackage{subfig}
\usepackage{bm}
\usepackage{amsfonts}
\usepackage{amsmath} 
\usepackage{multirow}
\usepackage{todonotes}
\usepackage{algorithm}
\usepackage{algorithmic}
\usepackage{setspace}
\usepackage{mathtools}

\DeclarePairedDelimiter\floor{\lfloor}{\rfloor}
\begin{document}
\title{Enhanced Uncertainty Estimation in Ultrasound Image Segmentation with MSU-Net}
\titlerunning{Uncertainty Estimation with MSU-Net}
\author{Rohini Banerjee* \and Cecilia G. Morales* \and
Artur Dubrawski}

\authorrunning{R. Banerjee et al.}

\institute{Carnegie Mellon University, Pittsburgh PA 15213, USA \email{\{rohinib,cgmorale,awd\}@andrew.cmu.edu}}

\maketitle              % typeset the header of the contribution
\def\thefootnote{*}\footnotetext{These authors contributed equally to this work}\def\thefootnote{\arabic{footnote}}

\begin{abstract}%\todo{cut paper to 8 pages content + 2 pages cite}
Efficient intravascular access in trauma and critical care significantly impacts patient outcomes. However, the availability of skilled medical personnel in austere environments is often limited. Autonomous robotic ultrasound systems can aid in needle insertion for medication delivery and support non-experts in such tasks. Despite advances in autonomous needle insertion, inaccuracies in vessel segmentation predictions pose risks. Understanding the uncertainty of predictive models in ultrasound imaging is crucial for assessing their reliability. We introduce MSU-Net, a novel multistage approach for training an ensemble of U-Nets to yield accurate ultrasound image segmentation maps. We demonstrate substantial improvements, 18.1\% over a single Monte Carlo U-Net, enhancing uncertainty evaluations, model transparency, and trustworthiness. By highlighting areas of model certainty, MSU-Net can guide safe needle insertions, empowering non-experts to accomplish such tasks.

\end{abstract}
\section{Introduction}

Trauma, the leading cause of death among young individuals in the U.S.~\cite{wallace2023fluid}, often results in blood loss, which requires rapid fluid resuscitation for vital organ oxygenation. In austere settings, accessing timely medical care can be challenging due to limited access, dangerous conditions, time constraints, or the absence of medical infrastructure, making expertise for optimal needle insertion sites critical. Autonomous robotic systems can assist in intravenous fluid administration when medical experts are unavailable, providing support in emergencies. These systems can also guide non-experts in accurate performance of phlebotomy tasks, empowering them to contribute effectively in dire medical situations.

Ultrasound imaging is widely used for locating vessels to enable fluid resuscitation due to its affordability, speed, safety, and portability, unlike CT or MRI imaging that are not portable and use ionizing radiation~\cite{GoelMotionawareNS}. Despite advancements in autonomous needle insertion into blood vessels~\cite{GoelMotionawareNS}, a critical challenge persists: inaccurate predictions can lead to severe consequences. Failing to anticipate vessel structure during intravenous cannulation may result in catastrophic outcomes for critically injured individuals; incorrectly predicting two adjacent vessels as a single one could lead to laceration of the vessel wall and cause hemorrhage upon needle insertion~\cite{morales2024bifurcation}. Both medical personnel and automated tools need to assess the certainty of their estimates of vessel location and structure to mitigate this risk~\cite{morales20233d}. 
As healthcare increasingly adopts automation, reliance on Artificial Intelligence (AI) models grows. Building a  model that communicates its uncertainty becomes essential to help users focus their attention and actions on where the model is confident, such as guiding needle insertion accurately within vessel segments. In this paper, we attack the problem of uncertainty estimation of models trained to guide needle insertion tasks. Our contributions include: 
(1) We introduce MSU-Net, a novel Multistaged Monte Carlo U-Net, demonstrating significant advancements in uncertainty quantification; (2) Our model achieves state-of-the-art performance in ultrasound image segmentation, marking the first known improvement in uncertainty estimation for this application.

\section{Related Works}
\subsection{Uncertainty Quantification in Deep Learning}
Uncertainty quantification is vital in understanding the reliability of AI model predictions. Traditionally, the frequentist approach assumes a single point estimate of network weights, using estimated class likelihoods as confidence measures for predictions. However, studies have shown that these likelihoods often overestimate accuracy~\cite{DBLP:journals/corr/GuoPSW17}, and the popular metric used to quantify confidence, expected calibration error, has been criticized for bias and inconsistency~\cite{gruber2024better}. This limits utility, motivating the need for alternative approaches to accurately quantify model uncertainty rather than just confidence.

Predictive uncertainty comprises of aleatoric and epistemic components~\cite{ghoshal-uncert}. Aleatoric uncertainty accounts for inherent noise in observations, while epistemic uncertainty arises from limited training data and model parameter uncertainty. Recent advancements in Bayesian inference and Bayesian neural networks have provided robust frameworks to quantify both forms of uncertainty by estimating posterior distributions over model weights. Gal and Ghahramani~\cite{gal2016dropout} introduced Monte Carlo (MC) dropout for Bayesian inference in deep learning, leveraging dropout in convolutional layers for stochastic forward passes to approximate Bayesian variational inference. Bayesian approximation using MC dropout has been extensively applied: Kendall and Gal~\cite{kendall2016bayesian} developed Bayesian SegNet for scene understanding, while Dechesne and Lassalle~\cite{bayesian-massachusetts} used it in U-Net for high-accuracy image segmentation. Seedat~\cite{seedat2020mcunet} proposed a human-in-the-loop system using model uncertainty. Yet, single-model architectures are now supplanted by model ensembles due to difficulties in capturing inherent variability.

\subsection{Multistage Neural Network Ensembles}
Training multiple individual models through model ensembling offers a range of tools to encourage member model diversity and improve overall accuracy. Diversity can be achieved through several techniques, the most popular being bagging~\cite{bagging}, stacking~\cite{wolpert-stacking}, and boosting~\cite{boosting}. Non-Bayesian deep ensembles improve predictive uncertainty estimation compared to single model architectures~\cite{lakshminarayanan2017simple}, yet remain limited by naïve aggregation strategies such as simple or weighted averaging or majority vote~\cite{yangbrowne}. Yang et al.~\cite{yang-og} addresses drawbacks of traditional ensembles by training a secondary neural network to adaptively assign weights, inspired by stacked generalization. This approach offers flexibility in candidate selection and leverages non-linear modeling capabilities of neural networks.

Lai et al.~\cite{lai-credit-risk-og} designed a multistage reliability-based neural network ensemble learning approach for credit card evaluation based on a decorrelation maximization procedure to select diverse models. Their approach combines outputs using different aggregation strategies, achieving superior accuracy over single and hybrid models. Multistage ensembling with U-Nets also proves beneficial; Yin and Hu~\cite{pawnet} designed Paw-Net, a multistage ensemble for semantic segmentation. Paw-Net demonstrates higher final Intersection-over-Union (IoU) scores by integrating outputs of multiple U-Nets specialized in different classes. 

Our proposed MSU-Net architecture combines the benefits of these appro\-aches: candidate U-Nets are first overproduced and then selected based on decorrelation to compose an ensemble. The final stage combines outputs using a Monte Carlo U-Net approach to produce accurate segmentation maps.

\section{Methods}
\subsection{Monte Carlo U-Net (MCU-Net)}
We introduce stochasticity into our inference process by incorporating dropout layers into our chosen U-Net architecture, thereby enabling MC dropout, as in MCU-Net ~\cite{seedat2020mcunet}. MCU-Net is henceforth referred to as our baseline model. Indiscriminately placing dropout after each convolutional layer can lead to reduced fit and poor test performance~\cite{kendall2016bayesian}. We opt to situate dropout layers in the decoder section of our U-Net instead. Each decoder block consists of two sets of convolutional layers (3x3 convolution filter, batch normalization, ReLU activation) and an attention block. This enables us to approximate Bayesian inference by conducting $T$ forward passes, or MC samples, of the U-Net during testing. Our empirical results show no significant improvement beyond $T=30$. Our model outputs both logits and logit variances~\cite{chen-galeotti-ub} to use for segmentation and epistemic uncertainty maps,
% \begin{equation}
% [\widehat{p}_t, \widehat{\sigma}^2_t] = f^{\widehat{\omega}_{t}}(x)
% \end{equation}
with  $[\widehat{p}_t, \widehat{\sigma}^2_t] =f^{\widehat{\omega}_{t}}(x)$ representing the $t$-th forward pass of MCU-Net with learned weights $\widehat{\omega}_t$. Given sigmoid activation, $\bm{\sigma}_{\text{SIG}}$, the ensemble prediction is  aggregated by averaging individual model outputs, $\bm{\sigma}_{\text{SIG}}(\widehat{p}_t)$,  across all $T$ samples. Similarly, the raw epistemic uncertainty map is obtained by averaging logit variances across all MC samples.

\subsection{Multistage U-Net (MSU-Net)}
\begin{figure}
\vspace{-5pt}
\begin{center}
    \includegraphics[scale=.33]{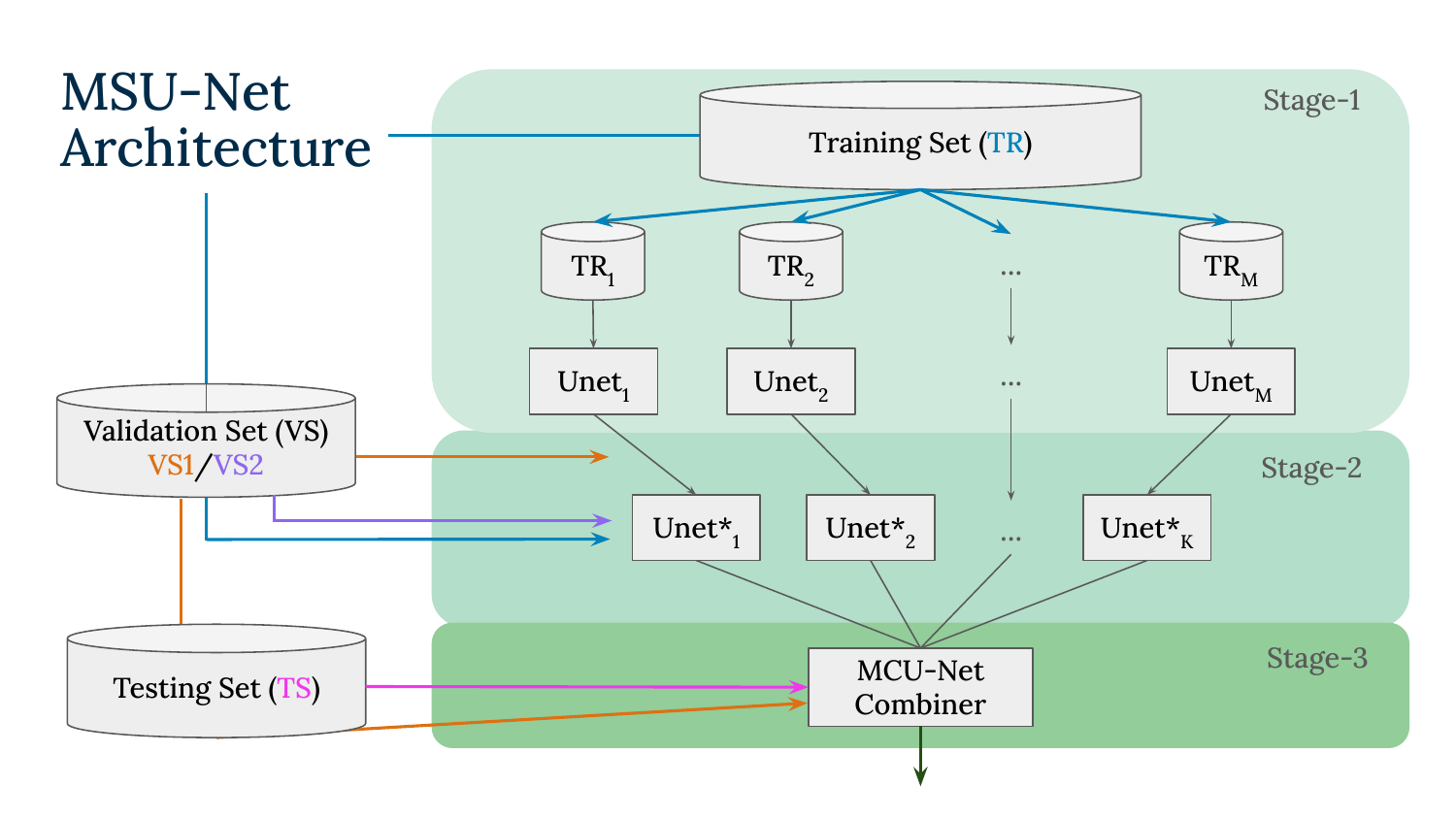}
\end{center}
\vspace{-20pt}
\caption{Proposed MSU-Net architecture. U-Nets are trained on bootstrap samples and validated on VS1. Decorrelated ensemble members are chosen using VS2.} \label{msun_arch}
\vspace{-15pt}
\end{figure}

The MSU-Net architecture, illustrated in Fig~\ref{msun_arch}, is structured in three main stages: (1) ensembling, (2) candidate selection, and (3) combining. In stage 1, we train multiple models referred to as \textit{candidates}. Bagging is employed to induce diversity into our training data; given the training dataset of size $n_{TR}$, we repeatedly sample with replacement to generate $M$ training subsets, each also of size $n_{TR}$. We then proceed to train a total of 15 bootstrapped models~\cite{lakshminarayanan2017simple}. Hyperparameter optimization, specifically determining the number of epochs of training for each candidate model, is performed using the VS1 validation set, held out for this purpose. Early stopping is applied to each candidate when validation loss begins to increase, to mitigate model overfitting. 

In stage 2, candidates are selected to form a diverse and efficient ensemble, aiming to reduce computational costs by minimizing correlation. We apply the decorrelation maximization method from~\cite{lai-credit-risk-og}, using the Brier score loss matrix computed from models $f^{\widehat{\omega}_1}_1, f^{\widehat{\omega}_2}_2, \cdots, f^{\widehat{\omega}_M}_M$ on a separate validation set VS2 within the specified region of interest, replacing binary error with Brier score for finer error measurement. The mean, variance, and covariance of the Brier score loss matrix is then computed in order to construct the correlation matrix $R$, where the $i$th row and $j$th column represents the degree of correlation between $f^{\widehat{\omega}_i}_i$ and $f^{\widehat{\omega}_j}_j$. We write $R$ in a block form for candidate model $f^{\widehat{\omega}_i}_i$ as

\begin{equation}
    R \xrightarrow{\text{extract principle submatrix} }\begin{bmatrix}
        R_{-i} & r_i \\ r_i^T & 1
    \end{bmatrix}
\end{equation}

\noindent where $R_{-i}$ is the principle submatrix of $R$ resulting from deleting the $i$th row and $i$th column. Subsequently, we compute the plural-correlation coefficient~\cite{oneill2021multiple}
\begin{equation}
    \rho_i^2 = r_i^T R^{-1}_{-i} r_i
\end{equation}
by which candidate $f^{\widehat{\omega}_i}_i$ is kept if $\rho_i^2 \leq \theta$ for some threshold $\theta$, else discarded, in order to build our final ensemble. Empirical trials testing $\theta$ thresholds indicate that $K=3$ performs almost equally as well as $K=15$, suggesting that minimal additional training will suffice for our new architecture.

% \begin{algorithm}
% \caption{Decorrelation maximization}\label{alg:decor_max}
% \begin{algorithmic}[1]
%     \STATE \textbf{Input:} Candidate models $(f_1, f_2, \cdots, f_M)$, correlation matrix $R$, threshold $\theta$.
%     \STATE \textbf{Output:} Chosen ensemble members $(f^*_1, f^*_2, \cdots, f^*_K)$
%     \STATE $\texttt{ensemble} \gets \texttt{[]}$
%     \FOR{$f_i$ in $(f_1, f_2, \cdots, f_M)$}
%         \STATE Rewrite $R$ in block matrix form for candidate model $f_i$, i.e. $R \rightarrow \begin{bmatrix}
%         R_{-i} & r_i \\ r_i^T & 1
%         \end{bmatrix}$
%         \STATE $\rho_i^2 \gets r_i^T R^{-1}_{-i} r_i$
%         \IF{$\rho_i^2 \leq \theta$}
%             \STATE $\texttt{ensemble}$ += $[f_i]$
%         \ELSE
%             \STATE Discard $f_i$
%         \ENDIF
%     \ENDFOR
%     \RETURN \texttt{ensemble}
% \end{algorithmic}
% \end{algorithm}

In stage 3, we train a final MCU-Net combiner taking ensemble member outputs as inputs and outputting segmentation maps. We compute the segmentation and epistemic uncertainty maps of MSU-Net analogous to MCU-Net.
\vspace*{-6pt}
\section{Experiments}
\vspace*{-6pt}
\noindent{\bf Dataset and Training \ }
We used a ultrasound scanning system described by Morales et al.~\cite{Morales2023}  
to scan the CAE Blue Phantom anthropomorphic gel model simulating femoral vessels. Equipped with a 5MHz linear transducer, the system can scan up to 5cm in depth, producing 2D transverse ultrasound images. Expert clinicians annotated these images using the Computer Vision Annotating Tool (CVAT)~\cite{boris_sekachev_2020_4009388}, followed by cropping and resizing to 256 $\times$ 256 pixels. The dataset was split into training (1392 images), validation (907 images), and testing (856 images). The validation set was further randomly split into two disjoint sets, VS1 and VS2.

For image segmentation, we employed U-Net architectures with a ResNet34 backbone in Pytorch, utilizing the Segmentation Models library~\cite{Iakubovskii:2019} pretrained on ImageNet. To integrate MC Dropout, dropout layers with optimal rates of 0.4 and 0.5 were added after each ReLU activation  in the decoder~\cite{kendall2016bayesian,kim2023use}. Training used a batch size of 8, Adam optimizer with a learning rate of 0.0001, and early stopping based on validation loss stabilization. MSU-Net incorporated bagging for ensembling and the plural-correlation coefficient as a correlation metric.

% The final additional selected hyperparameters of each architecture are shown in Table~\ref{hyperparameters}.

% \begin{table}
% \caption{Selected hyperparameters of each model architecture.}\label{hyperparameters}
% \begin{center}
% \begin{tabular}{|l|l|c|c|}
% \hline
% \multirow{2}{*}{\bf{Task}} & \multirow{2}{*}{\bf{Hyperparameters}} &
%     \multicolumn{2}{c|}{\bf{Value}} \\
% \cline{3-4}
%  &  & \multicolumn{1}{c|}{MCU-Net} & \multicolumn{1}{c|}{MSU-Net} \\
% %\cline{2-5}
% \hline
% Model architecture & Ensemble size & $-$ & 15 \\
% & Ensemble technique & $-$ & Bagging \\ 
% & Correlation metric & $-$ & Plural-correlation coefficient \\
% \hline
% Model training & Early stopping epoch & 15 & 150\\
% \hline
% \end{tabular}
% \end{center}
% \end{table}

\begin{figure}
    \vspace{-15pt}
    \begin{center}
        \includegraphics[scale=.26]{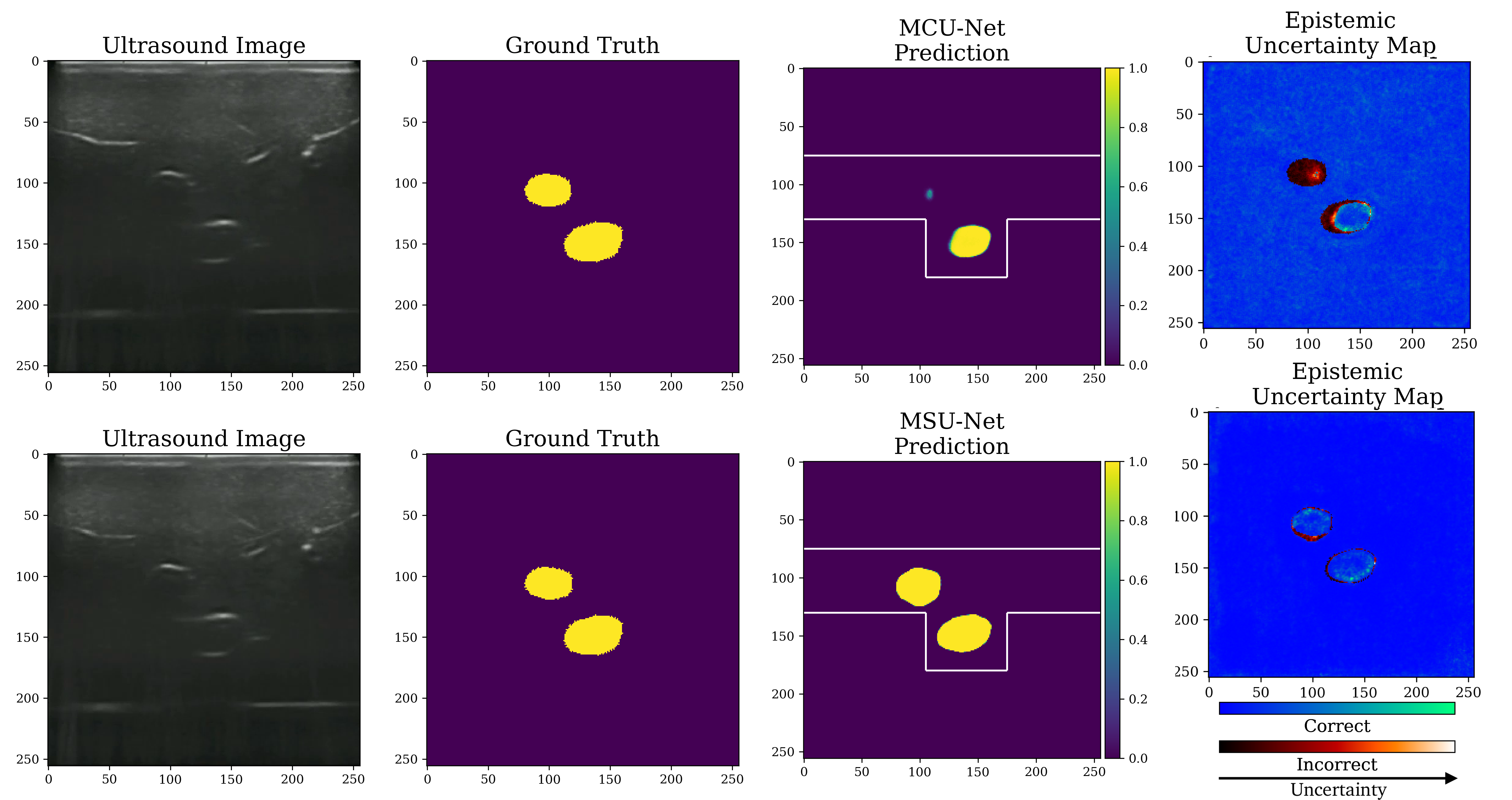}
    \end{center}
    \vspace{-15pt}
    \caption{Epistemic uncertainty maps on test data for (a) MCU-Net and (b) MSU-Net. Darker/lighter colors indicate lower/higher uncertainty values. To alleviate class imbalance, we limit evaluations to a region of interest delineated in white.} \label{qual-maps}
    \vspace{-15pt}
\end{figure}

% \begin{figure}%
%     \vspace{-20pt}
%     \centering
%     \subfloat{{\includegraphics[scale=0.33]{figures/mcunet_ex.png}}}%
%     \vspace{1em}
%     \subfloat{{\includegraphics[trim={0 0 0 0.07em},clip, scale=0.33]{figures/msunet_ex.png}}}
%     \caption{Epistemic uncertainty maps on test data for (a) MCU-Net and (b) MSU-Net. Darker/lighter colors indicate lower/higher uncertainty values. To alleviate class imbalance, we limit evaluations to a region of interest delineated in white.}%
%     \label{qual-maps}%
%     \vspace{-20pt}
% \end{figure}

\noindent{\bf Distribution Divergence Estimation}
Model quality hinges on uncertainty distribution: low for correct, high for incorrect predictions, enhancing calibration and user trust~\cite{shamsi2023uncertaintyaware}. We utilize the Rényi divergence (RD) statistic as our metric for comparison. It is a generalization of Kullback-Leibler (KL) divergence that measures the dissimilarity between two probability distributions $p$ and $q$. van Erven and Harremoës~\cite{van_Erven_2014} relay an operational characterization of RD as the number of bits by which a mixture of two codes, $p$ and $q$, can be compressed.  For some measurable set $\mathcal{M}_0$ that $p$ and $q$ lie in and order $\alpha \in \mathbb{R} \setminus \{1 \}$, RD of distribution $p$ from distribution $q$ is defined as:

\begin{equation}
R_{\alpha}(p\,||\,q) = \frac{1}{\alpha-1} \ln \int_{\mathcal{M}_0} p^{\alpha}(x) q^{1-\alpha}(x) \,dx
\end{equation}

A nonparametric estimator of RD that is conditionally $L_2$-consistent using only $k$-nearest-neighbor statistics has been proposed to considerably reduce computational effort~\cite{pmlr-v15-poczos11a}. Let $X_{1:n_0}=(X_1, \cdots, X_{n_0})$ be an i.i.d. sample from a distribution with density $p$ and $Y_{1:n_1}=(Y_1, \cdots, Y_{n_1})$  an i.i.d. sample from a distribution with density $q$. We denote $\rho_k(i)$ to be the $k$-th nearest neighbor of observation $X_i$ in $X_{1:{n_0}}$ and $v_k(i)$ the $k$-th nearest neighbor of $X_i$ in $Y_{1:{n_1}}$. With $B_{k, \alpha} = \frac{\Gamma(k)^2}{\Gamma(k-\alpha+1) \Gamma(k+\alpha-1)}$, we can estimate the RD by:

\begin{equation}
\widehat{R}_{\alpha}(p\,||\,q) = \frac{1}{\alpha-1} \ln \left( n_0^{-1} \sum_{i=1}^{n_0} \left(\frac{(n_0-1) \rho_k(i)}{n_1 v_k(i)}\right)^{1-\alpha} B_{k,\alpha}\right)
\end{equation}

We use RD to measure the ability of our method to distinguish correct predictions ($p$) from incorrect predictions ($q$). We conduct permutation testing to assess deterministicity and bootstrapping to obtain confidence intervals for the results. Nearest neighbor estimators are sensitive to perturbations in the underlying distribution, and hence their limited variance cannot be consistently estimated by a naïve Efron-type bootstrap~\cite{abadie-failure}. %since it induces samples with a higher positive probability of ties than the original sample distribution for large sample sizes. 
Since this behavior may result in a non-negligible positive bias in bootstrap estimates, we instead apply a direct M-out-of-N (MooN) type bootstrap~\cite{walsh2023} for this metric, shown in Algorithm~\ref{alg:moon}.

\vspace*{-6pt}
\begin{algorithm}
\caption{M-out-of-N bootstrapping}\label{alg:moon}
\begin{spacing}{0.9}
\begin{algorithmic}[1]
    \STATE \textbf{Input:} Distributions $p$, $q$. Degree of undersampling, $\gamma \in (0,1]$.
    \STATE \textbf{Output:} Bootstrap estimates of nonparametric Rényi divergence statistic
    \STATE $n_0, n_1 \gets |p|, |q|$
    \STATE $\alpha_n \gets \frac{n_0}{n_1}$
    \STATE $N^* \gets \floor*{(n_0 + n_1)^\gamma + \frac{1}{2}}$
    \STATE $n^*_0 \gets \floor*{\frac{\alpha_n}{1+\alpha_n} N^* + \frac{1}{2}}$
    \STATE $n^*_1 \gets N^* - n^*_0$
    \STATE \texttt{samples} $\gets$ \texttt{[]}
    \FOR{$i$ in range $1000$}
        \STATE \texttt{boot$\_$p} $\gets$ \texttt{resample($p$, n\_samples=$n^*_0$)} with replacement
        \STATE \texttt{boot$\_$q} $\gets$ \texttt{resample($q$, n\_samples=$n^*_1$)} with replacement
        \STATE \texttt{samples} $\stackrel{+}{=}$ \texttt{[}$\widehat{R}_{\alpha=0.85}$(\texttt{boot$\_$p}, \texttt{boot$\_$q})\texttt{]}
    \ENDFOR
    \RETURN \texttt{samples}
\end{algorithmic}
\end{spacing}
\end{algorithm}
\vspace*{-18pt}

\section{Results and Discussion}

\begin{figure}[h]
\begin{center}
    \includegraphics[scale=.4]{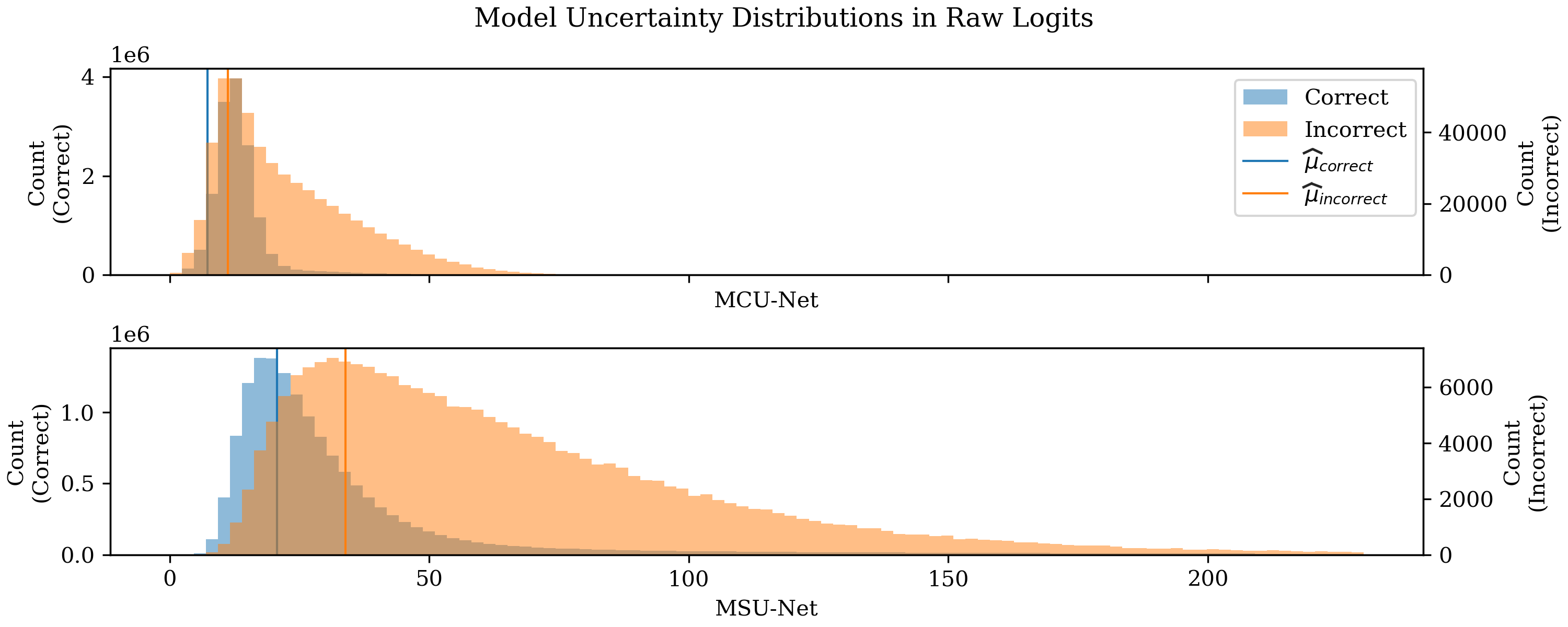}
\end{center}
\vspace{-15pt}
\caption{Epistemic uncertainty distributions for correct (blue) and incorrect (orange) predictions for MCU-Net (top) and MSU-Net (bottom). Our approach yields a markedly better differentiation of correct and incorrect predictions.} \label{uncert_dist}
\end{figure}

We evaluate MSU-Net using quantitative and qualitative metrics. To alleviate heavy class imbalance, we choose to average metrics over a predefined region of interest (ROI), shown in  Fig.~\ref{qual-maps}. Fig.~\ref{uncert_dist} displays model uncertainty distributions on the test set. We employ kernel density estimation with a Gaussian kernel and optimal bandwidth via Silverman’s rule of thumb to visualize difference in means between correct and incorrect prediction uncertainty distributions over 100,000 samples. To quantify separation between these distributions with RD, we select $k=4$ for $k$-nearest neighbors and $\alpha=0.85$ for numerical stability and interpretability~\cite{pmlr-v15-poczos11a}. We perform $B=1000$ permutations and bootstrapped samples. For MooN-type bootstrap, $\gamma=0.8$ is selected to maintain the largest proportion of original data while achieving the closest coverage probability of $0.95$ for $95\%$ intervals. Our final results are displayed in Table~\ref{model-quality-res}.

\begin{table}[h]
\vspace{-20pt}
\caption{Model quality in ROI. Arrows show direction of better performance.}\label{model-quality-res}
\begin{center}
\begin{tabular}{|c|c|c|c|c|c|c|}
\hline
 &  $\widehat{\mu}_{\textit{corr}}$& $\widehat{\mu}_{\textit{incorr}}$ & $\Delta \widehat{\mu}(\uparrow)$ & $\widehat{R}_{\alpha}(corr\,||\,incorr)(\uparrow)$ & $95\%$ CI on $\widehat{R_\alpha}$ & $p$-value $(\downarrow)$\\
\hline
 MCU-Net & 7.230 & 11.229 & 3.999 & 0.429 & [0.426, 0.453]  & 0.003 \\
 MSU-Net & 20.783 & 33.876 & \textbf{13.093} & \textbf{0.638} & [0.603, 0.667] & 0.003 \\
\hline
\end{tabular}
\end{center}
\vspace{-20pt}
\end{table}

\begin{figure}%
\vspace{0pt}
    \begin{center}
        \includegraphics[scale=.33]{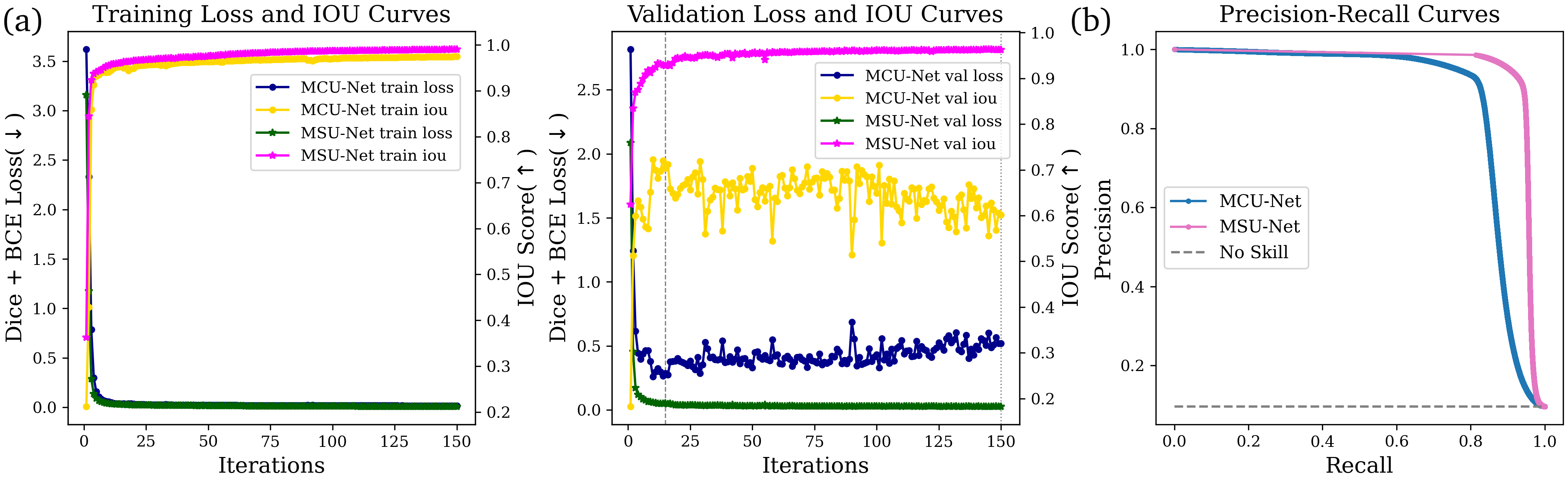}
    \end{center}
    \vspace{-10pt}
    \caption{(a) Training (left) and validation (right) curves for MCU-Net and MSU-Net. Early stopping delineated by gray lines. (b) Precision-recall curves.}%
    \label{model-performance}%
\vspace{-15pt}
\end{figure}

Fig.~\ref{model-performance}a shows training behavior for both models. MSU-Net shows a more stable convergence and consistently outperforms MCU-Net during validation performed after each epoch. MSU-Net achieves $18.1\%$ better mean IoU and a significant improvement in sensitivity and false negative rate scores at alpha level $0.05$, while other metrics remain similar. We additionally utilize precision-recall curves, seen in  Fig.~\ref{model-performance}b, which are resilient to unbalanced classes since they only focus on positive class predictions. Performance results are displayed in Table~\ref{model-perf-res}.

\begin{table}
\vspace{-20pt}
\caption{Model performance in ROI.}\label{model-perf-res}
\begin{center}
\begin{tabular}{|c|c|c|c|c|c|}
\hline
 & Mean test IoU($\uparrow$) & Specificity$(\uparrow)$& Sensitivity$(\uparrow)$ & FPR$(\downarrow)$ & FNR$(\downarrow)$ \\
\hline
 MCU-Net & 0.679 & \textbf{0.998} & 0.673 & \textbf{0.002} & 0.327 \\
 MSU-Net & \textbf{0.860} & 0.996 & \textbf{0.890} & 0.004 & \textbf{0.110} \\
\hline
\end{tabular}
\end{center}
\vspace{-20pt}
\end{table}

\vspace*{-18pt}
\vspace*{-12pt}

Permutation tests for MCU-Net and MSU-Net ($p \le 0.003$ for both) indicate that the observed separation between correct and incorrect predictions is not random in both. Yet, at the $95\%$ confidence level, we see no overlap between their $95\%$ confidence intervals, showing that the ability of MSU-Net to distinguish correct from incorrect predictions is significantly better than that of MCU-Net.

Qualitative uncertainty maps visually validate our findings and capture local variations in model performance~\cite{bayesian-massachusetts}. MCU-Net exhibits indiscriminately low uncertainty in Fig.~\ref{qual-maps}, whereas MSU-Net provides more interpretable uncertainty values. MSU-Net shows clearer vessel tops with lower uncertainty and higher uncertanties at the bottom of vessels which is consistent with our assessment. 
MSU-Net yields a higher credibility in assessing its predictive reliability. 

MSU-Net generally outperforms MCU-Net. Although it performs marginally worse at specificity and false positive rate (FPR), the precision-recall curves from Fig.~\ref{model-performance}b show a higher average precision score for MSU-Net at 0.945 than MCU-Net at 0.875 compared to the baseline score of 0.09 for a `no-skill' classifier. As such, at the same level of recall, MSU-Net correctly classifies a greater proportion of pixels that are actually vessels than MCU-Net. Crucially, MSU-Net achieves a considerably lower false negative rate (FNR) than MCU-Net. In this context, failing to anticipate a true vessel can have more catastrophic consequences to a critically injured individual than incorrectly anticipating a vessel. MSU-Net improves credibility not only through higher quality results, but also through more accurate results while avoiding potentially disastrous deficiencies of predictive modeling in the context of medical image segmentation.

\section{Conclusion}
This paper introduces MSU-Net, a multistaged Monte Carlo U-Net for uncertain ultrasound image segmentations. It improves model transparency and trustworthiness compared to standard U-Net by enhancing uncertainty evaluation, despite minimal additional training. Our framework sets a benchmark for future studies, offering qualitative maps for intuitive model assessment. While our preliminary results are limited to phantom data and binary image segmentation, we aim to validate these findings on live animal and human data in future work.

\begin{credits}
\subsubsection{\ackname} Authors thank Nico Zevallos, Dr.\ Michael R.\ Pinsky, and Dr.\ Hernando Gomez for gathering experiment data. This work was partially supported by the U.S.\ Dept.\ of Defense contracts W81XWH-19-C0083 and W81XWH-19-C0101.

\subsubsection{\discintname}
The authors have no competing interests to declare that are
relevant to the content of this article.
\end{credits}

\newpage
 \bibliographystyle{splncs04}
 \bibliography{bibliography}




% LNCS Homepage, \url{http://www.springer.com/lncs}, last accessed 2023/10/25
% \end{thebibliography}
\end{document}